\documentclass{article}
\usepackage{kvpath,times}
\newcommand{\ReferenceStyle}{plainnat}

\usepackage{amsmath,amsfonts,bm}

\def\eqref#1{equation~\ref{#1}}

\def\1{\bm{1}}

\DeclareMathAlphabet{\mathsfit}{\encodingdefault}{\sfdefault}{m}{sl}
\SetMathAlphabet{\mathsfit}{bold}{\encodingdefault}{\sfdefault}{bx}{n}

\usepackage{hyperref}
\usepackage{url}
\usepackage{amsmath,amssymb,bm,booktabs}
\usepackage{tikz}
\usepackage{pgfplots}
\usepgfplotslibrary{groupplots}
\usepackage{multirow,multicol}
\usetikzlibrary{arrows.meta,positioning,fit,calc}
\usepackage{graphicx}

\newcommand{\nop}[1]{}

\title{KVPath: Connecting Key-Value Representations Across Transformer Layers}
\author{Xuan Luo, Zhiyuan Wang, Hanzhong Cao, Xifeng Yan \\
University of California, Santa Barbara \\
{\normalfont\small\texttt{xuan\_luo@ucsb.edu}, \texttt{zwang796@ucsb.edu}} \\
{\normalfont\small\texttt{hcao5@umail.ucsb.edu}, \texttt{xyan@cs.ucsb.edu}}
}

\hypersetup{pdftitle={KVPath: Connecting Key-Value Representations Across Transformer Layers},pdfauthor={Xuan Luo, Zhiyuan Wang, Hanzhong Cao, Xifeng Yan},pdfsubject={},pdfkeywords={}}

\begin{document}
\maketitle

\newcommand{\huggingface}{\includegraphics[height=1.4em]{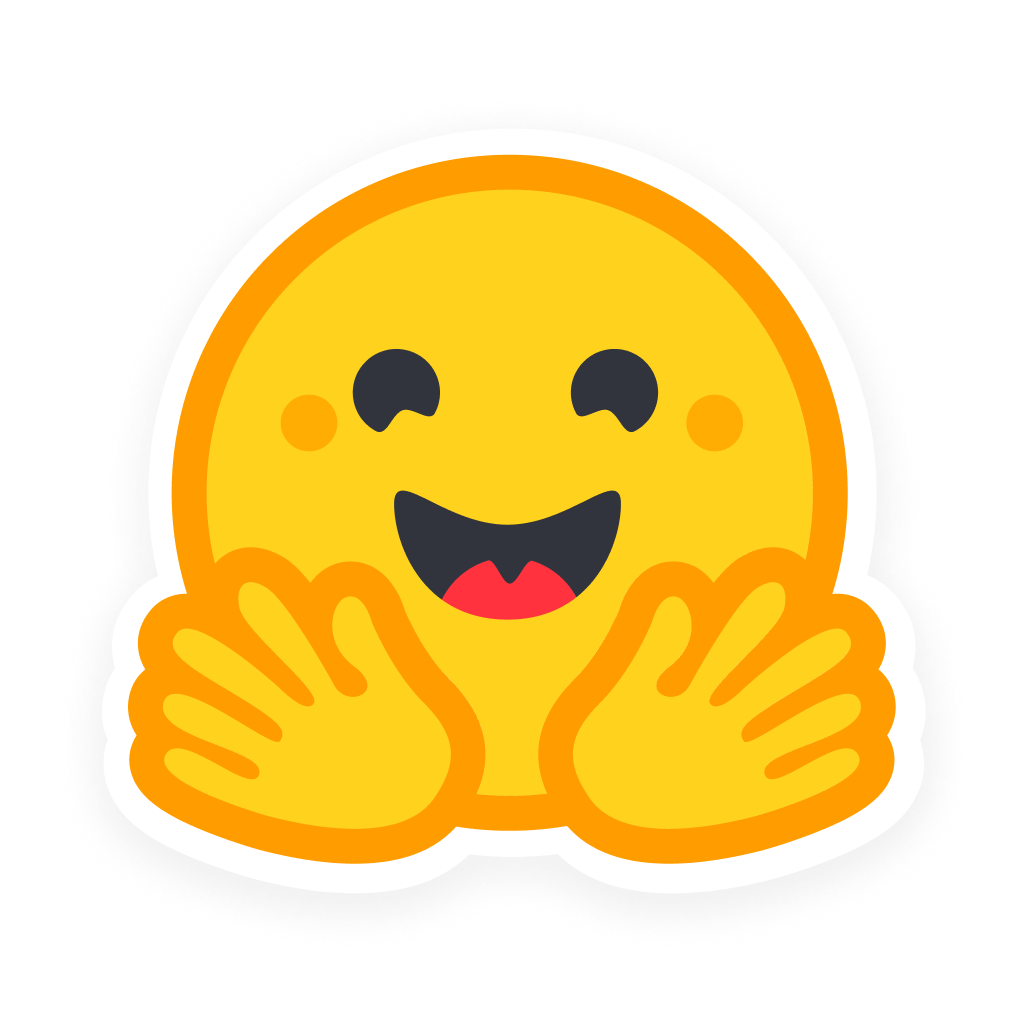}}
\newcommand{\github}{\includegraphics[height=1.2em]{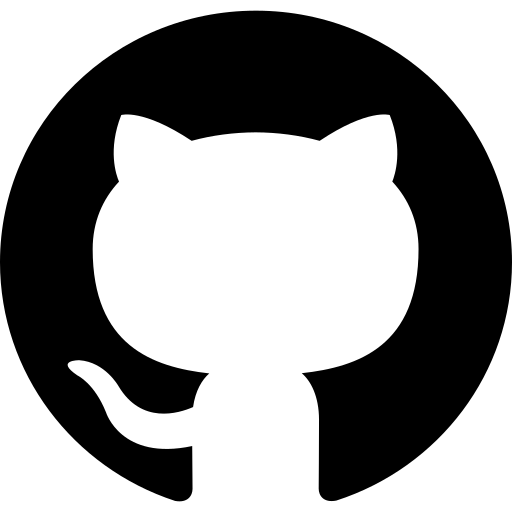}}

\begin{abstract}
Transformers typically cache an independent key-value (KV) representation for each token at every layer. Recent work reveals that a token's KV representations across neighboring layers are correlated and contain shared information. We hypothesize that this correlation implies incremental inter-layer changes, which can be effectively captured by low-dimensional deltas. Motivated by this, we propose KVPath, a new KV representation paradigm that connects KV representations across the network depth into a path. The path establishes a full-dimensional base KV at the first layer, while subsequent layers generate low-dimensional delta KVs to update the preceding KV, thereby constructing their own KV states. By caching the base KV once alongside these compact delta KVs, KVPath preserves distinct layer-specific representations without storing full KV states at every layer. We develop two variants of KVPath: KVPath-Residual, which updates vector values via residual addition, and KVPath-Rolling, which progressively updates feature dimensions in a rolling manner. Under matched parameter counts, KVPath reduces KV-cache storage by up to 68\% for Grouped-Query Attention (GQA) and 55\% for Multi-Head Latent Attention (MLA), while maintaining or even improving model performance.

\begin{center}
\begin{tabular}{lcp{0.55\textwidth}}
\github & \href{https://github.com/luoxuan-cs/KVPath}{\color{cyan}\path{https://github.com/luoxuan-cs/KVPath}} \\
\huggingface & \href{https://huggingface.co/xuan-luo/KVPath}{\color{cyan}\path{https://huggingface.co/xuan-luo/KVPath}} \\
\end{tabular}
\end{center}

\end{abstract}

\section{Introduction}

Large language models (LLMs)~\citep{attn,gpt2,gpt3,palm} rely on the key-value (KV)~\citep{kv} cache to avoid recomputing the key-value representations of previous tokens during autoregressive generation. For each token, a standard Transformer independently produces and stores key and value representations at every layer. Consequently, KV-cache memory grows linearly with both sequence length and network depth, making it a major bottleneck for long-context inference and high-throughput serving~\citep{vllm,sglang}. While substantial effort has been devoted to reducing KV-cache cost through quantization~\citep{kivi}, token eviction~\citep{h2o}, and low-dimensional representations~\citep{palu}, most approaches treat the KV representations produced at different Transformer layers independently.

Recent work has begun to exploit cross-layer correlation in KV representations to reduce the KV-cache footprint. Architectures such as CLA~\citep{cla} and YOCO~\citep{yoco} reuse KV across multiple layers, while SwiftKV~\citep{swiftkv} adapts pretrained models to cross-layer KV sharing through distillation. Additionally, MiniCache~\citep{minicache} merges neighboring layers' KV while retaining distinct entries, and xKV~\citep{xkv} jointly factorizes multiple layers' KV into a shared low-rank subspace. These studies demonstrate that KV representations across network depth are correlated and contain overlapping information.

To better leverage this cross-layer redundancy, we look to how Transformers naturally process representations across depth. Rather than building each layer's representation independently, Transformers compute hidden states through a continuous trajectory:
\begin{equation}
h^{(0)}
\xrightarrow{\Delta h^{(1)}} h^{(1)}
\xrightarrow{\Delta h^{(2)}} \cdots
\xrightarrow{\Delta h^{(L-1)}} h^{(L-1)},
\end{equation}
where each transition follows $h^{(\ell)} = h^{(\ell-1)} + \Delta h^{(\ell)}$ for $1 \leq \ell < L$. Inspired by this depth-wise dynamics, rather than viewing cross-layer KV redundancy as static shared copies or isolated approximations, we can model their relationship as a continuous, depth-wise trajectory. Since keys and values are directly derived from hidden states at each layer, it is natural to ask: can KV representations similarly form a path across depth through progressive layer-to-layer updates? Just as hidden states advance via layer-wise residual additions, KV states can be similarly updated across depth. We further hypothesize that these layer-to-layer KV updates can be effectively captured by low-dimensional delta vectors.

In this paper, we introduce \textbf{KVPath}, a new KV representation paradigm that connects each token's KV across layers into a single path:
\begin{equation}
KV^{(0)}
\xrightarrow{\Delta KV^{(1)}} KV^{(1)}
\xrightarrow{\Delta KV^{(2)}} \cdots
\xrightarrow{\Delta KV^{(L-1)}} KV^{(L-1)}.
\end{equation}
Each path is encoded by a full-dimensional base KV and a sequence of low-dimensional delta KV updates. Specifically, the first layer generates the base KV, while each subsequent layer generates a low-dimensional delta KV from its current hidden state and updates the preceding layer's KV to derive its own state. We develop two variants of KVPath to instantiate these updates. KVPath-Residual up-projects a low-dimensional vector and accumulates it onto the preceding KV state for KV updating. KVPath-Rolling divides the KV state into multiple sub-states and refreshes them in a rolling manner by continuously replacing the oldest sub-state with a new one. Both variants cache the base KV once together with these compact delta KVs along the path, preserving distinct layer-specific representations without storing full KV states at every layer. KVPath incorporates this path structure directly into the model architecture, with all parameters learned jointly during pre-training.

At matched model sizes and comparable model quality, KVPath reduces KV-cache storage by up to 68\% for Grouped-Query Attention (GQA) and 55\% for Multi-Head Latent Attention (MLA), while maintaining or even improving model performance. These results highlight that modeling inter-layer KV representations as a continuous path offers a promising perspective for efficient Transformer design, enabling more effective use of cache capacity.

\section{Method}
\label{sec:method}

We introduce KVPath, a KV representation paradigm that connects key and value states across network depth. Figure~\ref{fig:kvpath} illustrates two variants of KVPath: KVPath-Residual updates the full state through residual connections, whereas KVPath-Rolling progressively updates feature dimensions in a rolling manner. Both designs are learned end-to-end during pre-training.
\subsection{Path Representation}
\label{sec:kvpath_framework}

For each token and KV head, the states across all $L$ layers form
one \emph{path}, indexed by $\ell=0,\ldots,L-1$. The path is built from two types of components: a full-dimensional
\emph{base KV} and low-dimensional \emph{delta KV}.
The base KV, denoted by $(k^{(0)},v^{(0)})$, is generated at
the first layer ($\ell=0$) and initializes the path.
Each subsequent layer generates a delta KV, denoted by
$(k_\delta^{(\ell)},v_\delta^{(\ell)})$, to introduce new
information into the inherited state.
Both are projected from their respective layers'
hidden states:
\begin{equation}
\begin{aligned}
    (k^{(0)},v^{(0)}) &= W_0 h^{(0)},\\
    (k_\delta^{(\ell)},v_\delta^{(\ell)})
        &=W_\delta^{(\ell)}h^{(\ell)},
        \qquad 1\leq\ell<L,
\end{aligned}
\label{eq:kvpath_components}
\end{equation}
where $h^{(\ell)}\in\mathbb{R}^{D}$ is the normalized hidden state
input to the attention module of layer $\ell$, and $D$ is the
model's hidden-state dimension.
The base vectors $k^{(0)},v^{(0)}\in\mathbb{R}^{d}$ span the
full KV-head dimension $d$, while the delta vectors
$k_\delta^{(\ell)},v_\delta^{(\ell)}\in\mathbb{R}^{d_\delta}$
have a smaller dimension $d_\delta<d$.
Accordingly, $W_0\in\mathbb{R}^{2d\times D}$ and
$W_\delta^{(\ell)}\in\mathbb{R}^{2d_\delta\times D}$.
Token and KV-head indices are suppressed unless needed.

We use $(k^{(\ell)},v^{(\ell)})$ to denote the resulting
KV state at layer $\ell$, with
$k^{(\ell)},v^{(\ell)}\in\mathbb{R}^{d}$.
At the first layer, the KV state is initialized directly with the base KV.
Each subsequent layer combines the preceding layer's
KV state with its newly generated low-dimensional delta KV:
\begin{equation}
\begin{aligned}
    k^{(\ell)}
        &=\mathcal{F}_K^{(\ell)}
          \bigl(k^{(\ell-1)},k_\delta^{(\ell)}\bigr),\\
    v^{(\ell)}
        &=\mathcal{F}_V^{(\ell)}
          \bigl(v^{(\ell-1)},v_\delta^{(\ell)}\bigr),
    \qquad 1\leq\ell<L.
\end{aligned}
\label{eq:kvpath_general}
\end{equation}
The functions $\mathcal{F}_K^{(\ell)}$ and $\mathcal{F}_V^{(\ell)}$
specify how the inherited KV states are updated.
We next present two concrete implementations of these functions:
KVPath-Residual and KVPath-Rolling.

\subsection{Path Construction}
\label{sec:path_constructions}

\begin{figure}[t]
    \centering
    \includegraphics[width=\linewidth]{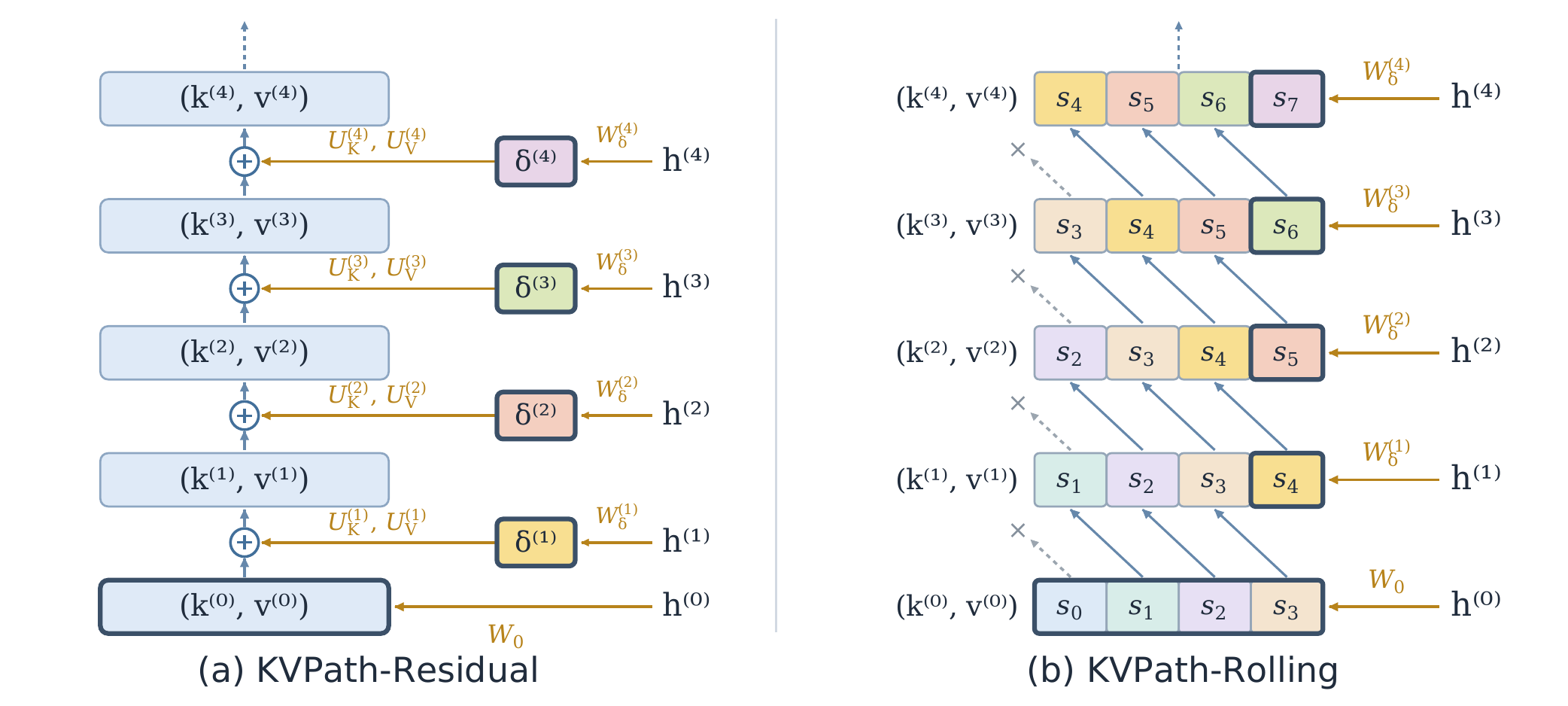}
    \caption{Two KVPath constructions across depth, shown from bottom
    to top. (a) KVPath-Residual adds an expanded low-dimensional
    update to the preceding full state.
    (b) KVPath-Rolling replaces the oldest slot
    with a new low-dimensional state.}
    \label{fig:kvpath}
\end{figure}

Figure~\ref{fig:kvpath} illustrates the two KVPath architectures:
KVPath-Residual and KVPath-Rolling.
Thick borders mark components newly generated from the current
hidden state: the full-dimensional base KV at layer $0$ and
the low-dimensional delta KV at subsequent layers.
Matching colors across panels identify new components generated
at corresponding layers. The figure focuses on raw KV states;
positional encoding and attention are described in
Section~\ref{sec:kvpath_attention}.

\paragraph{KVPath-Residual: cumulative low-dimensional updates.}
As shown in Figure~\ref{fig:kvpath}(a), KVPath-Residual starts
from the initial base state $(k^{(0)},v^{(0)})$ at layer $0$.
At each subsequent layer $\ell\geq1$, the network generates
a low-dimensional delta KV $\delta^{(l)}=(k_\delta^{(\ell)},v_\delta^{(\ell)})$
according to Eq.~\ref{eq:kvpath_components}.
It is then expanded via learned matrices
$U_K^{(\ell)},U_V^{(\ell)}\in\mathbb{R}^{d\times d_\delta}$
and added directly to the preceding layer's state.
The query is independently generated from the current hidden
state at every layer:
\begin{equation}
\begin{aligned}
    q^{(\ell)}
        &=W_Q^{(\ell)}h^{(\ell)},
        &&0\leq\ell<L,\\
    k^{(\ell)}
        &=k^{(\ell-1)}+U_K^{(\ell)}k_\delta^{(\ell)},\\
    v^{(\ell)}
        &=v^{(\ell-1)}+U_V^{(\ell)}v_\delta^{(\ell)},
        &&1\leq\ell<L.
\end{aligned}
\label{eq:residual_update}
\end{equation}
To cache a token, KVPath-Residual stores the full base KV
state once alongside the compact delta KV updates for the
subsequent $L-1$ layers, reducing the total cached
representation size per KV head from $2Ld$ to
$2d+2(L-1)d_\delta$.

\paragraph{KVPath-Rolling: a moving window of narrow states.}
As shown in Figure~\ref{fig:kvpath}(b), KVPath-Rolling applies
the rolling construction to queries, keys, and values.
It starts from the initial state
$(q^{(0)},k^{(0)},v^{(0)})$ at layer $0$, with each vector
partitioned into $m=d/d_\delta$ narrow slots of width
$d_\delta$:
\begin{equation}
\begin{aligned}
    q^{(0)}
        &=[s_0^Q\,\Vert\,\cdots\,\Vert\,s_{m-1}^Q],\\
    k^{(0)}
        &=[s_0^K\,\Vert\,\cdots\,\Vert\,s_{m-1}^K],\\
    v^{(0)}
        &=[s_0^V\,\Vert\,\cdots\,\Vert\,s_{m-1}^V],
\end{aligned}
\label{eq:rolling_initial}
\end{equation}
where $s_i^Q,s_i^K,s_i^V\in\mathbb{R}^{d_\delta}$ and
$\Vert$ denotes concatenation along the head-feature dimension. Here, $m$ denotes the number of slots in each layer's state,
while slot indices are assigned globally across layers.
As shown in Figure~\ref{fig:kvpath} (b), the initial slots in layer 0 have indices $0,\ldots,m-1$, and each
subsequent layer $\ell\geq1$ introduces one new slot for
each of Q/K/V with index $m+\ell-1$.

At each subsequent layer $\ell\geq1$, the network generates
a low-dimensional delta KV $(k_\delta^{(\ell)},v_\delta^{(\ell)})$
according to Eq.~\ref{eq:kvpath_components}, together with
a delta query
$q_\delta^{(\ell)}=W_{Q,\delta}^{(\ell)}h^{(\ell)}
\in\mathbb{R}^{d_\delta}$,
where $W_{Q,\delta}^{(\ell)}\in\mathbb{R}^{d_\delta\times D}$
is a learned projection that produces only the new query slot.
It removes the oldest slot from each state and appends
$s_{\ell+m-1}^Q=q_\delta^{(\ell)}$,
$s_{\ell+m-1}^K=k_\delta^{(\ell)}$, and
$s_{\ell+m-1}^V=v_\delta^{(\ell)}$, yielding
\begin{equation}
\begin{aligned}
    q^{(\ell)}
        &=[s_\ell^Q\,\Vert\,\cdots\,\Vert\,s_{\ell+m-1}^Q],\\
    k^{(\ell)}
        &=[s_\ell^K\,\Vert\,\cdots\,\Vert\,s_{\ell+m-1}^K],\\
    v^{(\ell)}
        &=[s_\ell^V\,\Vert\,\cdots\,\Vert\,s_{\ell+m-1}^V],
    \qquad 1\leq\ell<L.
\end{aligned}
\label{eq:rolling_state}
\end{equation}
The remaining slots are carried forward unchanged, so each
layer retains full $d$-dimensional Q/K/V states while
introducing only one new $d_\delta$-dimensional slot per state.

To cache a token, KVPath-Rolling stores the $m$ initial KV
slot pairs once alongside one new KV slot pair for each
subsequent layer, reducing the total cached representation
size per KV head from $2Ld$ to
$2(m+L-1)d_\delta=2d+2(L-1)d_\delta$.

\subsection{Attention with KVPath}
\label{sec:kvpath_attention}

To preserve fine-grained local information, we follow existing
practices in the literature
\citep{deepseekv4,deepseekv41,dual_dimensionality} by employing
a local attention window of size $w$.
To avoid redundant computations, we reuse the projections for local window attention
wherever possible: KVPath-Residual uses the same layer-specific
full-dimensional query for tokens inside and outside the window
($q^{(\ell)}=\bar{q}^{(\ell)}$), whereas KVPath-Rolling
slices a portion of the local full KV projections to serve
as the delta KV.

For a query--key pair separated by $r$ tokens, attention uses
the layer's full-dimensional local projections
$(\bar{q}^{(\ell)},\bar{k}^{(\ell)},\bar{v}^{(\ell)})$
when the key token is inside the window, and the KVPath
representations $(q^{(\ell)},k^{(\ell)},v^{(\ell)})$
when it is outside the window:
\begin{equation}
    (q,k,v)=
    \begin{cases}
        (\bar{q}^{(\ell)},\bar{k}^{(\ell)},\bar{v}^{(\ell)}),
            &0\leq r<w,\\
        (q^{(\ell)},k^{(\ell)},v^{(\ell)}),
            &r\ge w.
    \end{cases}
\label{eq:kvpath_window}
\end{equation}
Here, the window includes the current token ($r=0$).
For KVPath-Residual, the query is the same for tokens inside
and outside the window. A single softmax normalizes the attention scores over all
causal keys inside and outside the window, yielding the final
attention output as the weighted sum of the selected values.

KVPath constructs paths independently
for each KV head and preserves the mapping between query
heads and KV heads, making it directly applicable to
MHA~\citep{attn}, GQA~\citep{gqa}, and MQA~\citep{mqa}.
It also extends to MLA~\citep{mla} by constructing paths
over compressed KV latents, as described in the next subsection.

\subsection{KVPath on MLA}
\label{sec:mla-compatibility}

Multi-head latent attention (MLA)~\citep{mla} caches a
compressed KV latent shared across attention heads.
Each head reconstructs its own content keys and values
through head-specific up-projections.
Queries and keys also contain positional components,
with the positional key shared across heads and cached
separately. KVPath extends MLA by applying the path
constructions in Section~\ref{sec:path_constructions}
to the compressed latent space.

We denote the compressed latent dimension by $d_c$ and reuse
$d_\delta$ for the update dimension. The reconstructed
content keys and values have dimension $d_h$ per head,
while the positional components have dimension $d_R$.
Below, we suppress the head index: queries and
up-projections are head-specific, while latents and
updates are shared across heads.

Both ResMLA and RollMLA initialize the path with the
first layer's latent and reconstruct keys and values
at each layer:
\begin{equation}
\begin{aligned}
    c^{(0)}
        &=W_0 h^{(0)},\\
    (k^{(\ell)},v^{(\ell)})
        &=W_U^{(\ell)}
          \operatorname{RMSNorm}^{(\ell)}(c^{(\ell)}),\\
    k_R^{(\ell)}
        &=W_R^{(\ell)}h^{(\ell)}.
\end{aligned}
\label{eq:kvpath-mla}
\end{equation}
Here, $W_0 \in \mathbb{R}^{d_c \times D}$ is the initial projection matrix that maps the normalized hidden state of the first layer into the full-dimensional base compressed latent $c^{(0)}$. $W_U^{(\ell)} \in \mathbb{R}^{2d_h \times d_c}$ represents the head-specific up-projection matrix at layer $\ell$, which reconstructs the content keys and values from the layer's latent state after RMSNorm. Finally, $W_R^{(\ell)} \in \mathbb{R}^{d_R \times D}$ is the decoupled positional key projection matrix that derives the shared RoPE-based key component $k_R^{(\ell)}$ directly from the hidden state at each layer.

Each subsequent layer $\ell \ge 1$ obtains a compact latent update $\delta^{(\ell)} \in \mathbb{R}^{d_\delta}$ from its fresh hidden state via a learned projection $W_\delta^{(\ell)} \in \mathbb{R}^{d_\delta \times D}$:
\begin{equation}
    \delta^{(\ell)} = W_\delta^{(\ell)} h^{(\ell)}, \qquad 1 \le \ell < L.
\end{equation}
Conceptually, this update can be viewed as comprising compact key and value directional increments $\delta^{(\ell)} = (k_{\delta}^{(\ell)}, v_{\delta}^{(\ell)})$ embedded within the shared latent space. This update is then incorporated into the inherited state using one of the following constructions.

\paragraph{ResMLA.}
ResMLA expands the compact update to the full latent
dimension and adds it to the preceding state:
\begin{equation}
    c^{(\ell)}
        =c^{(\ell-1)}+U_c^{(\ell)}\delta^{(\ell)},
        \qquad 1\leq\ell<L,
\label{eq:resmla}
\end{equation}
where
$U_c^{(\ell)}\in\mathbb{R}^{d_c\times d_\delta}$
is a learned up-projection matrix. The latent accumulates
information continuously across all $L$ layers via residual additions.

\paragraph{RollMLA.}
RollMLA partitions the latent into $d_c/d_\delta$
slots. Each layer removes the oldest slot and appends
the current update:
\begin{equation}
    c^{(\ell)}
        =
        \left[
        c^{(\ell-1)}[d_\delta:d_c]
        \,\Vert\,\delta^{(\ell)}
        \right],
        \qquad 1\leq\ell<L,
\label{eq:rollmla}
\end{equation}
where $c^{(\ell-1)}[d_\delta:d_c]$ retains the final
$d_c-d_\delta$ dimensions of the preceding state, and $\Vert$ denotes concatenation along the feature dimension.
Thus, each layer carries forward the remaining slots
unchanged while introducing one new $d_\delta$-dimensional
slot $\delta^{(\ell)}$.

Both variants retain MLA's layer-specific queries and
positional keys. Following
Section~\ref{sec:kvpath_attention}, tokens within the local
window use keys and values reconstructed from the current
layer's fresh latent, while more distant tokens use those
reconstructed from $c^{(\ell)}$. Both branches reuse the
same layer-specific normalization and up-projection.

For either construction, the compact path stores the
initial latent, $L-1$ updates, and one positional key per
layer, requiring
$d_c+(L-1)d_\delta+Ld_R$ values per token.

\section{Experiments}

\subsection{Implementation Details}
\label{sec:implementation-details}

We train all the models on FineWeb-Edu~\citep{fineweb} using the GPT-2
BPE tokenizer~\citep{gpt2}. Training proceeds in three stages:
we first train from scratch at a context length of 2K on 10 billion tokens,
then extend the context to 8K and subsequently to 32K, using 1 billion
additional tokens at each stage. Each extension is initialized from the
final checkpoint of the preceding stage. We use Rotary Position Embeddings
(RoPE)~\citep{rope} with the RoPE base fixed at $10^6$ throughout training.
We use AdamW~\citep{adamw} with $\beta_1=0.9$, $\beta_2=0.95$,
$\epsilon=10^{-8}$, weight decay 0.1, and gradient clipping at 1.0.
The global batch sizes are 256, 64, and 16 sequences for the 2K, 8K,
and 32K stages, respectively. The peak learning rate is $10^{-3}$ for the initial
2K stage and $10^{-4}$ for the 8K and 32K extensions. We use linear
learning-rate warmup followed by cosine decay~\citep{loshchilov2017sgdr}
at each stage, with 100 warmup steps.

\begin{table}[t]
\centering
\setlength{\tabcolsep}{4pt}
\renewcommand{\arraystretch}{1.08}
\caption{Main Results. All models feature a head dimension of 64 and 98.71M parameters. 
ResGQA and RollGQA use matched base KV and delta KV dimensions. 
The KV cache size is expressed as a percentage of standard GQA's storage requirement at the same context length.}
\label{tab:main-results}
\begin{tabular}{@{}llccccc@{\hspace{12pt}}c@{}}
\toprule[1.2pt]
& & \multicolumn{5}{c}{\textbf{Perplexity} $\downarrow$} & \\
\cmidrule(lr){3-7}
\textbf{Context} & \textbf{Model}
& C4 & FineWeb-Edu & Wikipedia & PG-19 & Avg.
& \textbf{KV cache} $\downarrow$ \\
\midrule
2K & GQA
   & 35.67 & 21.94 & 32.56 & 33.58 & 30.94 & 100.00\% \\
   & CLA
   & 35.37 & 21.73 & 31.50 & 33.35 & 30.49 & 50.78\% \\
   & ResGQA
   & \textbf{34.80} & \textbf{21.36} & \textbf{31.31}
   & \textbf{32.72} & \textbf{30.05} & \textbf{32.81\%} \\
   & RollGQA
   & 35.68 & 22.08 & 32.03 & 33.84 & 30.91 & \textbf{32.81\%} \\
\midrule
8K & GQA
   & 34.42 & 21.14 & 30.82 & 31.10 & 29.37 & 100.00\% \\
   & CLA
   & 34.09 & 20.92 & 29.60 & 30.67 & 28.82 & 50.20\% \\
   & ResGQA
   & \textbf{33.53} & \textbf{20.56} & \textbf{29.51}
   & \textbf{30.23} & \textbf{28.46} & \textbf{31.64\%} \\
   & RollGQA
   & 34.39 & 21.26 & 30.08 & 31.49 & 29.31 & \textbf{31.64\%} \\
\midrule
32K & GQA
    & 34.15 & 20.95 & 30.38 & 30.68 & 29.04 & 100.00\% \\
    & CLA
    & 33.81 & 20.73 & 29.11 & 30.12 & 28.44 & 50.05\% \\
    & ResGQA
    & \textbf{33.25} & \textbf{20.36} & \textbf{29.00}
    & \textbf{29.51} & \textbf{28.03} & \textbf{31.35\%} \\
    & RollGQA
    & 34.12 & 21.06 & 29.55 & 30.83 & 28.89 & \textbf{31.35\%} \\
\bottomrule[1.2pt]
\end{tabular}
\end{table}

\subsection{Main Results}
\label{sec:main-results}
We evaluate KVPath on Grouped-Query Attention (GQA) by applying its residual update and rolling update variants to GQA, resulting in two models named ResGQA and RollGQA, respectively. Both variants are
benchmarked against standard GQA~\citep{gqa} and Cross-Layer Attention
(CLA)~\citep{cla}, with all models configured to 98,710,016 parameters.
To ensure a strictly equal parameter budget, we adjust the intermediate
dimensions of the SwiGLU layers across architectures.

All four models consist of 12 layers, a hidden dimension of 512,
8 query heads, 4 key-value (KV) heads, and a head dimension of 64.
Following established practices for preserving fine-grained local
representations alongside compressed global context
~\citep{deepseekv4,deepseekv41,dual_dimensionality}, both ResGQA
and RollGQA employ a 32-token local window. Within this window, each layer
performs standard attention using full-dimensional queries, keys, and values. Standard CLA does not include a local window. For a fair comparison,
we augment our CLA baseline with the same 32-token window, within which
each layer also performs vanilla full attention. Outside the local window,
CLA uses its original cross-layer KV-sharing mechanism. Detailed configurations for all four models are provided below:

\paragraph{GQA.}
Our GQA baseline generates and caches keys and values at every layer.
It uses a SwiGLU intermediate dimension of 2,048. With 12 layers,
4 KV heads, and a head dimension of 64, its cache contains
$2 \times 12 \times 4 \times 64 = 6,144$ elements per token.

\paragraph{CLA.}
Our CLA baseline uses a head dimension of 64, four KV heads, a group
size of two, and an added local window of 32 tokens. Within the local window, each layer performs full-QKV attention, while
tokens outside the window use standard CLA with cross-layer KV sharing. With six groups,
caching the shared K/V states requires
$2 \times 6 \times 4 \times 64 = 3,072$ elements per token.
At a sequence length of 32,768, the total cache storage is reduced by
49.95\% relative to GQA.

\paragraph{ResGQA.}
ResGQA uses a head dimension of 64, a delta KV
dimension of 16, and a local window of 32 tokens.
The first layer initializes the full-dimensional base KV.
Each subsequent layer generates separate 16-dimensional
delta K and delta V from its hidden state, projects them
to 64 dimensions, and adds them to the preceding K/V states.
Caching the base KV and subsequent delta KV requires
$2 \times 4 \times (64 + 11 \times 16) = 1,920$ elements per
token.
At a sequence length of 32,768, the total cache storage is
reduced by 68.65\% relative to GQA.

\paragraph{RollGQA.}
RollGQA uses a head dimension of 64, a delta KV
dimension of 16, and a local window of 32 tokens.
The first layer initializes the full-dimensional base KV.
Each subsequent layer retains the final 48 dimensions of the
preceding K/V states and appends 16-dimensional delta KV
generated from its own hidden state.
Caching the base KV and subsequent delta KV requires
$2 \times 4 \times (64 + 11 \times 16) = 1,920$ elements per
token.
At a sequence length of 32,768, the total cache storage is
reduced by 68.65\% relative to GQA.

We evaluate on C4~\citep{raffel2020t5},
FineWeb-Edu~\citep{fineweb},
Wikipedia~\citep{wikimedia_dumps},
and PG-19~\citep{rae2020compressive}
under 2K, 8K, and 32K context lengths.
At each context length, models use the same evaluation data
and sequence construction.
We evaluate approximately 10M tokens per benchmark and report
GPT-2 token-level perplexity.

Table~\ref{tab:main-results} reports perplexity and KV-cache
storage under matched parameter counts.
Relative to GQA, ResGQA achieves the same cache reductions as RollGQA,
with corresponding average perplexity reductions of
2.88\%, 3.10\%, and 3.48\%.
RollGQA reduces KV-cache storage
by 67.19\%, 68.36\%, and 68.65\% at 2K, 8K, and 32K
context lengths, with average perplexity reductions of
0.10\%, 0.20\%, and 0.52\%, respectively.
Overall, KVPath substantially reduces KV-cache storage
compared to standard GQA while maintaining or improving
language modeling performance across context lengths.

\begin{table}[t]
\centering
\setlength{\tabcolsep}{4pt}
\renewcommand{\arraystretch}{1.08}
\caption{MLA Results. All models have ~99.69M parameters.
ResMLA and RollMLA use matched base latent and delta
latent dimensions. KV cache is expressed as a percentage of standard
MLA's storage requirement at the same context length.}
\label{tab:mla-main-results}
\begin{tabular}{@{}llccccc@{\hspace{12pt}}c@{}}
\toprule[1.2pt]
& & \multicolumn{5}{c}{\textbf{Perplexity} $\downarrow$} & \\
\cmidrule(lr){3-7}
\textbf{Context} & \textbf{Model}
& C4 & FineWeb-Edu & Wikipedia & PG-19 & Avg.
& \textbf{KV cache} $\downarrow$ \\
\midrule
2K & MLA
    & 36.27 & 22.22 & \textbf{32.22} & 34.88 & 31.40 & 100.00\% \\
    & CLLA
    & 35.76 & 22.22 & 32.59 & 34.71 & 31.32 & 52.47\% \\
    & ResMLA
    & \textbf{35.08} & \textbf{21.61} & 32.27
    & 34.50 & \textbf{30.87} & \textbf{46.56\%} \\
    & RollMLA
    & 35.51 & 21.87 & 32.67
    & \textbf{34.36} & 31.10 & \textbf{46.56\%} \\
\midrule
8K & MLA
    & 34.91 & 21.37 & 30.48 & 32.18 & 29.74 & 100.00\% \\
    & CLLA
    & 34.45 & 21.37 & 30.55 & 32.14 & 29.63 & 51.30\% \\
    & ResMLA
    & \textbf{33.80} & \textbf{20.80} & \textbf{30.44}
    & 32.27 & \textbf{29.33} & \textbf{45.39\%} \\
    & RollMLA
    & 34.18 & 21.03 & 30.68
    & \textbf{31.91} & 29.45 & \textbf{45.39\%} \\
\midrule
32K & MLA
    & 34.63 & 21.15 & 30.08 & 31.28 & 29.28 & 100.00\% \\
    & CLLA
    & 34.14 & 21.13 & 30.03 & 31.45 & 29.19 & 51.01\% \\
    & ResMLA
    & \textbf{33.51} & \textbf{20.60} & \textbf{29.99}
    & 31.63 & \textbf{28.93} & \textbf{45.10\%} \\
    & RollMLA
    & 33.90 & 20.81 & 30.22
    & \textbf{31.24} & 29.04 & \textbf{45.10\%} \\
\bottomrule[1.2pt]
\end{tabular}
\end{table}

\begin{table}[t]
\centering
\setlength{\tabcolsep}{4pt}
\renewcommand{\arraystretch}{1.08}
\caption{Scaling Results. Both models have 606.41M parameters
and are trained on approximately 56B tokens.
KV cache includes the local-window overhead and is expressed
as a percentage of standard GQA's storage requirement
at the same context length.}
\label{tab:scaling-ppl}
\begin{tabular}{@{}llccccc@{\hspace{12pt}}c@{}}
\toprule[1.2pt]
& & \multicolumn{5}{c}{\textbf{Perplexity} $\downarrow$} & \\
\cmidrule(lr){3-7}
\textbf{Context} & \textbf{Model}
& C4 & FineWeb-Edu & Wikipedia & PG-19 & Avg.
& \textbf{KV cache} $\downarrow$ \\
\midrule
2K & GQA
   & 21.93 & 12.92 & 19.63 & 20.29 & 18.69 & 100.00\% \\
   & RollGQA
   & \textbf{21.79} & \textbf{12.84} & \textbf{18.77}
   & \textbf{20.01} & \textbf{18.35} & \textbf{33.59\%} \\
\midrule
8K & GQA
   & 21.15 & 12.46 & 18.59 & 19.11 & 17.83 & 100.00\% \\
   & RollGQA
   & \textbf{21.02} & \textbf{12.38} & \textbf{17.73}
   & \textbf{18.52} & \textbf{17.41} & \textbf{28.91\%} \\
\midrule
32K & GQA
    & 21.06 & 12.39 & 18.40 & 19.06 & 17.73 & 100.00\% \\
    & RollGQA
    & \textbf{20.94} & \textbf{12.30} & \textbf{17.58}
    & \textbf{18.36} & \textbf{17.29} & \textbf{27.73\%} \\
\bottomrule[1.2pt]
\end{tabular}
\end{table}

\begin{table}[t]
\centering
\setlength{\tabcolsep}{6pt}
\renewcommand{\arraystretch}{1.08}
\caption{Zero-shot downstream evaluation at the 0.6B scale
after approximately 56B training tokens with a 2K context limit.
Both models have 606.41M parameters.
Accuracy scores are reported as percentages.}
\label{tab:scaling-benchmarks}
\begin{tabular}{@{}llcc@{}}
\toprule[1.2pt]
& & \multicolumn{2}{c}{\textbf{Model}} \\
\cmidrule(lr){3-4}
\textbf{Task} & \textbf{Metric}
& \textbf{GQA} & \textbf{RollGQA} \\
\midrule
HellaSwag
    & Acc.\ norm.\ ($\uparrow$)
    & 45.78 & \textbf{46.41} \\
PIQA
    & Acc.\ ($\uparrow$)
    & \textbf{69.42} & 68.61 \\
ARC-Easy
    & Acc.\ norm.\ ($\uparrow$)
    & 57.83 & \textbf{60.10} \\
WinoGrande
    & Acc.\ ($\uparrow$)
    & \textbf{55.25} & 52.33 \\
SciQ
    & Acc.\ ($\uparrow$)
    & 83.40 & \textbf{85.30} \\
LAMBADA
    & Acc.\ ($\uparrow$)
    & 36.81 & \textbf{38.15} \\
    & Target-word PPL ($\downarrow$)
    & 28.97 & \textbf{25.79} \\
BoolQ
    & Acc.\ ($\uparrow$)
    & 56.67 & \textbf{59.57} \\
OpenBookQA
    & Acc.\ norm.\ ($\uparrow$)
    & 34.00 & \textbf{35.40} \\
\bottomrule[1.2pt]
\end{tabular}
\end{table}

\subsection{Results with MLA}
\label{sec:mla-results}

We evaluate KVPath on Multi-Head Latent Attention (MLA)~\citep{mla}
by applying its residual update and rolling update variants to MLA, resulting in two models named ResMLA and RollMLA, respectively.
All MLA experiments use 12 layers, a hidden size of 512, 8 query heads,
a content-latent dimension of 128, and a decoupled RoPE key dimension
of 32. Queries are generated independently at every layer, and the
content component of each key and value has a per-head dimension of 64.
Both ResMLA and RollMLA employ a local window of 32 tokens, within which
each layer uses independently generated full-dimensional content
latents; tokens outside the window use the corresponding KVPath latent
representations. All models are matched to ~99.69M parameters by
adjusting their SwiGLU intermediate dimensions and follow the same
training and evaluation protocols as the GQA experiments.

\paragraph{MLA Baseline.}
The MLA baseline independently generates and caches a 128-dimensional
content latent and a 32-dimensional decoupled RoPE key at every layer. With 12 layers, its cache contains
$12 \times (128 + 32) = 1,920$ elements per token.

\paragraph{CLLA Baseline.}
We also include a Cross-Layer Latent Attention (CLLA) baseline under the same parameter budget. CLLA groups every two layers together to share a compressed KV latent of dimension 128, augmented with a 32-dimensional decoupled RoPE key per layer and a 32-token local attention window. At a sequence length of 32,768, CLLA reduces cache storage by approximately 48.99\% relative to standard MLA.

\paragraph{ResMLA.}
ResMLA uses a delta latent dimension of 32 and a local window
of 32 tokens. The first layer initializes the 128-dimensional base
latent. Each subsequent layer generates a 32-dimensional delta latent
from its hidden state, projects it to 128 dimensions, and adds it to
the preceding latent. Caching the base latent, eleven delta latents,
and twelve 32-dimensional RoPE keys requires
$128 + 11 \times 32 + 12 \times 32 = 864$ elements per token.
At a sequence length of 32,768, including the local-window overhead,
the total cache storage is reduced by 54.90\% relative to standard MLA.

\paragraph{RollMLA.}
RollMLA uses a delta latent dimension of 32 and a local window
of 32 tokens. The first layer initializes the 128-dimensional base
latent. Each subsequent layer retains the final 96 dimensions of the
preceding latent and appends a newly generated 32-dimensional delta
latent. Caching the base latent, eleven delta latents, and twelve
32-dimensional RoPE keys requires
$128 + 11 \times 32 + 12 \times 32 = 864$ elements per token.
At a sequence length of 32,768, including the local-window overhead,
the total cache storage is reduced by 54.90\% relative to standard MLA.

Table~\ref{tab:mla-main-results} reports perplexity and KV-cache
storage under matched parameter counts. Relative to standard MLA,
both ResMLA and RollMLA reduce KV-cache storage
by 53.44\%, 54.61\%, and 54.90\% at 2K, 8K, and 32K context
lengths, respectively. ResMLA achieves average perplexity reductions of
1.69\%, 1.38\%, and 1.20\%, while RollMLA achieves average perplexity reductions of
0.96\%, 0.98\%, and 0.82\%. Furthermore, both KVPath variants outperform the CLLA baseline while achieving even higher cache compression, demonstrating the effectiveness of continuous inter-layer path modeling over static group-sharing strategies.

\subsection{Scaling Analysis}
\label{sec:scaling}

We evaluate RollGQA against standard GQA at the 0.6B scale.
Both models have 606,405,632 parameters and consist of
32 layers, a hidden dimension of 1024, 16 query heads,
8 KV heads, a head dimension of 64, and a SwiGLU
intermediate dimension of 4096.
RollGQA uses four rolling slots of width 16 and a local attention window of 128 tokens.
Due to limited computational resources, we evaluate only
RollGQA at this scale.

We train both models on approximately 56B tokens:
50B at a context length of 2K, followed by 5B at 8K
and 1B at 32K, with global batch sizes of 1024, 256,
and 64 sequences, respectively.
Each stage is initialized from the final checkpoint
of the preceding stage.
All evaluations use the final 32K checkpoints.

Table~\ref{tab:scaling-ppl} reports perplexity and KV-cache
storage under matched parameter counts.
RollGQA achieves lower perplexity on all four benchmarks
across all three context lengths, with average perplexity
reductions of 1.83\%, 2.32\%, and 2.45\% at 2K, 8K,
and 32K, respectively.
At 32K, it reduces KV-cache storage by 72.27\% relative
to GQA.
We also evaluate both models on eight downstream tasks
in a zero-shot setting,
using identical evaluation examples and prompts.
As shown in Table~\ref{tab:scaling-benchmarks},
RollGQA achieves higher accuracy on six of the eight tasks
and reduces LAMBADA target-word perplexity from 28.97
to 25.79, while GQA performs better on PIQA and WinoGrande.

\subsection{Ablation Studies}
\label{sec:ablation-studies}

We examine the effect of the delta KV dimension on ResGQA
and RollGQA by varying $d_\delta \in \{8,16,32\}$.
All models use a head dimension of 64 and a local window of 32 tokens.
We maintain the same parameter count of 98,710,016 by adjusting the
SwiGLU intermediate dimensions where necessary. To summarize performance across context lengths, we average each
benchmark's perplexity over the 2K, 8K, and 32K evaluations.

\begin{table}[t]
\centering
\setlength{\tabcolsep}{4pt}
\renewcommand{\arraystretch}{1.08}
\caption{Ablation on the delta KV dimension $d$ with a 32-token local
window. All models have 98.71M parameters. Perplexity is averaged over
2K, 8K, and 32K evaluation contexts. KV cache is the mean percentage of GQA's storage
requirement across the same context lengths.}
\label{tab:local32-dimension-ablation}
\begin{tabular}{@{}lcccccc@{\hspace{12pt}}c@{}}
\toprule[1.2pt]
& & \multicolumn{5}{c}{\textbf{Perplexity} $\downarrow$} & \\
\cmidrule(lr){3-7}
\textbf{Model} & $d_\delta$
& C4 & FineWeb-Edu & Wikipedia & PG-19 & Avg.
& \textbf{KV cache} $\downarrow$ \\
\midrule
ResGQA & 8 & 33.91 & 20.74 & 29.47 & 30.99 & 28.78 & 20.48\% \\
 & 16 & 33.86 & 20.76 & 29.94 & 30.82 & 28.85 & 31.93\% \\
 & 32 & 33.85 & 20.77 & 29.28 & 30.47 & 28.59 & 54.85\% \\
\midrule
RollGQA & 8 & 35.63 & 21.99 & 31.27 & 32.54 & 30.36 & 20.48\% \\
 & 16 & 34.73 & 21.47 & 30.55 & 32.05 & 29.70 & 31.93\% \\
 & 32 & 34.84 & 21.50 & 32.59 & 32.79 & 30.43 & 54.85\% \\
\bottomrule[1.2pt]
\end{tabular}
\end{table}

Table~\ref{tab:local32-dimension-ablation} reports the results.
As $d_\delta$ increases from 8 to 32, the mean KV-cache storage increases
from 20.48\% to 54.85\% of GQA's storage requirement.
However, perplexity does not vary strictly monotonically with $d_\delta$. One possible explanation is that changing the delta KV dimension
affects several aspects of the model simultaneously, including
layer-specific update capacity and the reuse of representations
across layers.

\section{Limitations}
\label{sec:limitations}

Our experiments are limited to models with up to approximately
0.6B parameters due to computational constraints.
Whether the observed quality and KV-cache benefits persist
at larger model scales remains to be evaluated. We have not developed specialized attention kernels for
KVPath with complicated optimizations
available in systems such as FlashAttention~\citep{flashattention}
and FlashInfer~\citep{flashinfer}.
Efficiently accessing shared components and reconstructing
layer-specific states from the compact cache requires
nontrivial kernel design.
We therefore report compact KV-cache requirements rather
than throughput comparisons. Developing optimized kernels and evaluating end-to-end
serving performance are left for future work.

\section{Related Work}
\label{sec:related-work}

\paragraph{Within-layer KV-cache reduction.}
MQA~\citep{mqa} and GQA~\citep{gqa} reduce KV-cache storage
by sharing keys and values across query heads.
MLA~\citep{mla} caches compressed latents from which
head-specific keys and values are reconstructed.
Palu~\citep{palu} compresses KV representations through
low-rank projections.
LRKV~\citep{lrkv} combines shared KV representations with
low-rank head-specific residuals.
Other approaches reduce cache precision through
quantization~\citep{kivi} or retain fewer tokens through
eviction~\citep{h2o}.
These methods primarily address representation and retention
within individual layers, whereas KVPath models how KV
representations evolve across network depth.

\paragraph{Cross-layer KV-cache reduction.}
CLA~\citep{cla} shares KV within groups of layers,
YOCO~\citep{yoco} lets upper layers attend to a common KV cache,
and MLKV~\citep{mlkv} extends sharing across both heads and layers.
CLLA~\citep{clla} reconstructs layer-specific KV from shared latents.
FusedKV~\citep{fusedkv} constructs upper-layer KV through
layer-specific fusion of lower-layer sources.
YOCO++~\citep{yocopp} strengthens information flow by combining
bottom-layer KV with the KV generated by lower-half layers.
Other methods compress existing cross-layer representations:
MiniCache~\citep{minicache} merges neighboring layers' KV states,
xKV~\citep{xkv} jointly factorizes KV across layers,
and CommonKV~\citep{commonkv} combines joint weight decomposition
with latent-cache merging.
A related sequential-update idea appears in
XQuant-CL~\citep{xquant}, which encodes inter-layer differences in hidden states $X$ via quantization, progressively reconstructs these activations and rematerializes KV through layer-specific projections.

KVPath instead takes a path-based view of cross-layer KV representations:
each layer's KV is a state reached by updating the preceding
layer's state.
Rather than reusing or reconstructing KV from shared sources,
or jointly compressing independently generated states,
KVPath learns how KV evolves through successive layers.

\section{Conclusion}
\label{sec:conclusion}

We introduced KVPath, which models a token's KV representations
as an evolving state across network depth.
Rather than independently generating each layer's KV or
reusing the same KV across layers, KVPath constructs each
layer's state by updating the preceding state with compact
new information from the current hidden state.
Our residual and rolling constructions instantiate this
principle through cumulative updates and progressive slot
replacement, respectively.
Experiments with GQA and MLA show that these learned paths
substantially reduce KV-cache storage while maintaining or
improving model quality under matched parameter counts.
KVPath thus frames cross-layer KV efficiency as a problem
of learning how representations evolve across depth,
with compact updates serving as the mechanism for
layer-wise adaptation.

\section*{Acknowledgments}
We would like to thank Meta for the A100 GPUs used in our experiments. We also gratefully acknowledge the generous support of the NVIDIA Academic Grant Program and NCSA DeltaAI through allocation CIS260864 from the Advanced Cyberinfrastructure Coordination Ecosystem: Services \& Support (ACCESS) program, which is supported by U.S. National Science Foundation.

\subsection*{AI use statement}

We used generative AI tools to assist with language editing
and improve the clarity and readability of the manuscript.
We also used OpenAI Codex to assist with the implementation
of the model code used in this work.
The authors take responsibility for the final manuscript,
the implementation, and the reported results.

\bibliography{references}
\bibliographystyle{\ReferenceStyle}

\end{document}